\documentclass[10pt,twocolumn,letterpaper]{article}

\usepackage[pagenumbers]{wacv} 

\usepackage{times}
\usepackage{epsfig}
\usepackage{amsmath}
\usepackage{amssymb}
\usepackage{graphicx}
\usepackage{subcaption}
\usepackage{booktabs}
\usepackage{multirow}
\usepackage{bm}

\usepackage{listings}
\usepackage{xcolor}
\usepackage{todonotes}
\usepackage{subcaption}

\usepackage{algorithm}
\usepackage{algpseudocode}

\usepackage{pifont}
\newcommand{\rulesep}{\unskip\ \vrule\ }

\usepackage{array}
\newcommand{\PreserveBackslash}[1]{\let\temp=\\#1\let\\=\temp}
\newcolumntype{C}[1]{>{\PreserveBackslash\centering}p{#1}}

\definecolor{codegreen}{rgb}{0,0.6,0}
\definecolor{codegray}{rgb}{0.5,0.5,0.5}
\definecolor{codepurple}{rgb}{0.58,0,0.82}
\definecolor{codeblue}{rgb}{0.0,0.80,0.80}
\definecolor{codeorange}{rgb}{0.99,0.55,0.0}
\definecolor{backcolour}{rgb}{0.94,0.94,0.94}

\lstdefinestyle{mystyle}{
    backgroundcolor=\color{backcolour},   
    commentstyle=\color{codeorange},
    keywordstyle=\color{blue},
    numberstyle=\tiny\color{codegray},
    stringstyle=\color{codeblue},
    basicstyle=\ttfamily\footnotesize,
    breakatwhitespace=false,         
    breaklines=true,                 
    captionpos=b,                    
    keepspaces=true,                 
    numbers=left,                    
    numbersep=5pt,                  
    showspaces=false,                
    showstringspaces=false,
    showtabs=false,                  
    tabsize=2
}

\usepackage[pagebackref=true,breaklinks=true,letterpaper=true,colorlinks,bookmarks=false]{hyperref}

\def\wacvPaperID{1297} 
\def\confName{WACV}
\def\confYear{2024}

\begin{document}

\title{Attention-Guided Masked Autoencoders For Learning Image Representations}

\author{Leon Sick\\
Ulm University\\
{\tt\small leon.sick@uni-ulm.de}
\and
Dominik Engel\\
Ulm University\\
{\tt\small dominik.engel@uni-ulm.de}
\and
Pedro Hermosilla\\
TU Vienna\\
{\tt\small phermosilla@cvl.tuwien.ac.at}
\and
Timo Ropinski\\
Ulm University\\
{\tt\small timo.ropinski@uni-ulm.de}
}
\maketitle

\begin{abstract}
   Masked autoencoders (MAEs) have established themselves as a powerful method for unsupervised pre-training for computer vision tasks. While vanilla MAEs put equal emphasis on reconstructing the individual parts of the image, we propose to inform the reconstruction process through an attention-guided loss function. By leveraging advances in unsupervised object discovery, we obtain an attention map of the scene which we employ in the loss function to put increased emphasis on reconstructing relevant objects, thus effectively incentivizing the model to learn more object-focused representations without compromising the established masking strategy. Our evaluations show that our pre-trained models learn better latent representations than the vanilla MAE, demonstrated by improved linear probing and k-NN classification results on several benchmarks while at the same time making ViTs more robust against varying backgrounds.
\end{abstract}

\vspace{-4mm}
\section{Introduction}
\label{sec:intro}

Despite their emergence as a promising alternative to CNNs in computer vision, transformers bring their own training challenges to the table: unstable training, high computational demand, as well as the need for large datasets to achieve competitive results~\cite{khan2022transformers}. Recent work has shown that self-supervised Vision Transformer (ViT) pre-training~\cite{he2022masked, zhou2021ibot, caron2021emerging, chen2021empirical, bao2021beit} can help mitigate some of these issues and improve their performance for various vision tasks. While there have been different pre-training strategies proposed, one idea that has gained much traction is masking portions of an input image and training a ViT to reconstruct the missing parts of the image. This task has been shown to be most effective in the setting of an MAE~\cite{he2022masked}, which reconstructs the image with equal emphasis on each individual part. 
\begin{figure}[!thb]
  \centering
   \includegraphics[width=\linewidth, page=1]{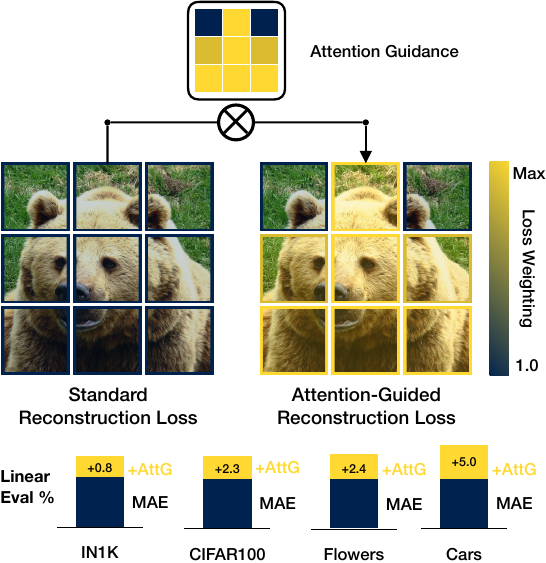}
   \caption{\textbf{Attention-Guided Reconstruction Loss.} We leverage an attention map, obtained from a self-supervised object discovery network, to inform our guided reconstruction loss \textbf{\emph{AttG}} towards reconstructing the relevant objects in the scene. Through this process, we guide the MAE to learn more effective off-the-shelf representations. 
}
   \label{fig:teaser}
   \vspace{-3mm}
\end{figure}
While MAEs achieve state-of-the-art representations for end-to-end finetuning tasks, it has also been shown that for linear evaluations, state-of-the-art results cannot be achieved~\cite{he2022masked, i-jepa}. 
With our work, we attempt to make progress towards closing this performance gap for MAEs on linear evaluations and k-NN classification. Methods from the field of explainability~\cite{selvaraju2017grad, simonyan2013deep, yosinskiunderstanding, chattopadhay2018grad} suggest that successful image classification models should mainly focus on the relevant objects in the scene to make their predictions. However, recent research has shown~\cite{Moayeri_2022_CVPR} that many failure cases in state-of-the-art models are due to the focus of these models on background elements.
Therefore, we hypothesize that inducing semantic knowledge into the MAE pre-training produces better latent representations, and we propose to induce information on the relevant object in the scene through an attention-guided reconstruction loss. While the vanilla MAE does not exhibit a notable deficiency in its capability to reconstruct objects, we show that guiding the model towards better reconstruction of the aforementioned allows it to learn representations more effective for linear evaluations. 
With our method, we aim to shift the process of learning representations more to the pre-training phase as to reduce the need for more expensive adaptation mechanisms like end-to-end finetuning. Our model therefore learns representations more powerful for evaluations that harness off-the-shelf features like linear probing and k-NN classification.
By manipulating the reconstruction loss, our model can be guided by an attention map to put increased emphasis on reconstructing relevant objects (see Fig.~\ref{fig:teaser}), which according to our ablations improves linear evaluations as compared to changing the established input masking strategy~\cite{li2022semmae, kakogeorgiou2022hide}. To generate the employed attention maps, we explore several object discovery methods, both ViT- and CNN-based. 
Our results show that our attention guidance mechanism \emph{AttG} narrows the gap for linear evaluation and k-NN classification tasks for MAEs. Thus, we obtain latent representations that enable increased top-1 accuracy for k-NN classification, where we see the biggest improvement, and linear probing in standard and few-shot settings. Furthermore, our learned, off-the-shelf representations prove to be especially useful on datasets not seen during pre-training as well as for image retrieval tasks. Further, by keeping a high input masking ratio, our method effectively enables our performance gains to be achieved without significantly increasing computational requirements.

\vspace{-1mm}
\section{Related Work}
\label{sec:formatting}

Our work builds upon recent advances in masked modeling, as well as self-supervised pre-training.

\noindent\textbf{Masked modeling.} Exploring a reconstruction task of masked regions in the input modality has first seen its successes in language processing with transformer architectures~\cite{vaswani2017attention, dosovitskiy2020image} in BERT~\cite{devlin2018bert} or GPT~\cite{brown2020language, radford2018improving, radford2019language}. More recently, these methods have been successfully transferred to the computer vision domain. In their work on masked autoencoders, He~\etal~\cite{he2022masked} mask a large number of image patches and feed the remaining tokens to the ViT~\cite{dosovitskiy2020image} encoder. Afterwards, the masked tokens are added and fed through the ViT decoder with the goal of reconstructing the masked patches. Feichtenhofer~\etal~\cite{feichtenhofer2022masked} and Tong~\etal~\cite{tong2022videomae} extend this idea into the spatio-temporal domain for video data, while Girdhar~\etal~\cite{girdhar2022omnimae} propose a unified MAE architecture for both image and video data.
Recently, masking strategies have also been applied for generative image transformers in work done by Chang~\etal~\cite{51195}. Their model learns to predict the initially masked tokens of the training image. To generate high-fidelity images at inference time, they sample the model to iteratively predict the tokens to be reconstructed and thereby refine the image.
Li~\etal~\cite{li2022semmae} propose to adapt a semantic-guided masking strategy for pre-training an MAE. They extract segmentation parts by refining the multi-head attention of a ViT encoder and use these to mask a portion or the whole part in the input image. While being effective, their approach requires additional finetuning on their segmentation model before extracting the segmentation parts. In contrast, our method does not require this preparation step and directly works with out of the box object discovery methods. Furthermore, we do not alter the original masking strategy, as it has proven to be beneficial by He~\etal~\cite{he2022masked}.
AttnMask, a novel masking strategy proposed by Kakogeorgiou et al. \cite{kakogeorgiou2022hide}, is integrated into a student-teacher knowledge distillation training process where the teacher network outputs an attention map that is used to mask the input image for the student network. They also extend their masking approach to keep some patches of the attention map visible, giving the model additional "hints". 
In their work on SimMIM, Xie~\etal~\cite{xie2022simmim} successfully design an approach for masked image modelling (MIM), for which they formulate a raw pixel regression task to predict the reconstruction of all masked pixels using a prediction head on top of the latent representation output by their encoder. Bao~\etal~\cite{bao2021beit} propose BEiT, which in contrast predicts masked patches in the token space, which is created from the image patches by a frozen discrete VAE.

\noindent\textbf{Self-supervised pre-training.} Recent work~\cite{tomasev2022pushing, caron2021emerging, caron2020unsupervised, chen2020improved, chen2020simple, mitrovic2020representation} has shown significant advances in self-supervised pre-training for vision models with various backbone architectures like ResNets~\cite{he2016deep} and ViTs~\cite{dosovitskiy2020image}. Chen~\etal~\cite{chen2020simple} explore pre-training for ResNets and highlight that effective data augmentation is a critical factor for pre-training on images. Their approach is formulated so that its performance is dependent on the batch size. Caron~\etal~\cite{caron2020unsupervised} built upon these findings and introduced a multi-crop augmentation strategy as well as a novel learning approach to eliminate the batch size dependence. Pre-training of ResNets is further pushed by works from Mitrovic~\etal~\cite{mitrovic2020representation} and Tomasev~\etal~\cite{tomasev2022pushing}. Their works propose to formulate a probabilistic learning objective through an invariance regularization mechanism, built upon the Kullback-Leibler divergence.
Research by Caron~\etal~\cite{caron2021emerging} puts special emphasis on exploring self-supervised pre-training with ViTs. They investigate the use of knowledge distillation and find that their ViTs can provide a segmentation of the scene and also perform well on k-NN classification. With MoCo~v3~\cite{chen2021empirical}, Chen~\etal also explore the properties of ViTs for pre-training in an evolution of their initial methods~\cite{he2020momentum, chen2020improved}.

\section{Method}
\begin{figure*}
  \centering
  \begin{subfigure}[t]{0.7\textwidth}
  \includegraphics[width=\textwidth, page=2]{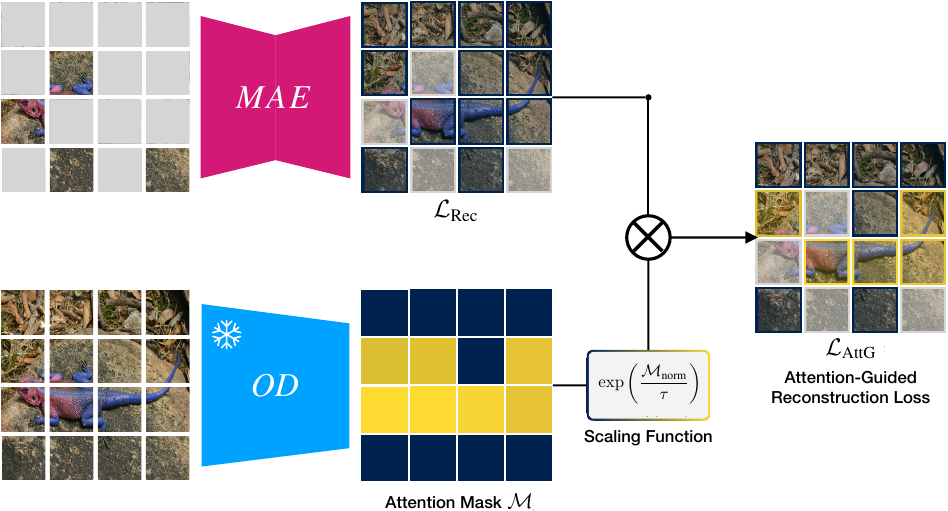}
  \caption{Attention-Guided Masked Autoencoder Architecture.}
  \label{subfig:main}
  \end{subfigure}%
    \hspace{0.5cm}
    \rulesep
    \hspace{0.5cm}
  \begin{subfigure}[t]{0.153\textwidth}
  \includegraphics[width=\textwidth, page=3]{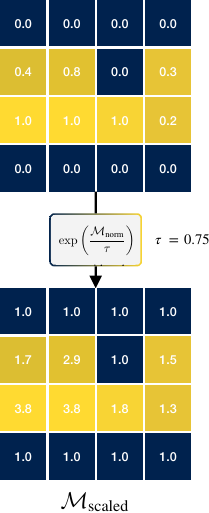}
  \caption{Scaling Function}
  \label{subfig:scaling}
  \end{subfigure}
  \caption{ \textbf{\emph{AttG} Architecture Overview And Scaling Function.}
  Our architecture, displayed in Figure~\ref{subfig:main}, employs two streams. Besides the main $MAE$ backbone, we facilitate an unsupervised object discovery ($OD$) stream with fixed weights to obtain a patch-based attention map. This attention map is then scaled and finally used to inform our attention guided reconstruction loss \emph{AttG} towards relevant objects. Gray overlays depict image patches which are not part of the loss since they have not been masked in the input image. To make the attention map useful for reconstruction guidance, we scale its values, as visualized in Figure~\ref{subfig:scaling}. This results in the loss for background patches not being masked out and patches from the relevant object in the scene being further emphasized.}
  \label{fig:main}
\end{figure*}

In our work, we build upon vanilla MAEs~\cite{he2022masked}. We enable them to learn more robust representations by steering their focus on objects of interest, without affecting the random patch sampling found to be essential by He~\etal~\cite{he2022masked}. As illustrated in Fig.~\ref{fig:main}, we facilitate unsupervised object discovery (OD) and use the obtained results within a temperature-controlled guidance loss to steer the model towards the relevant objects. In the following subsections, we will detail our learning architecture (Section~\ref{subsec:architecture}), introduce our attention-guided loss (Section~\ref{subsec:loss}), and discuss unsupervised OD (Section~\ref{subsec:objectdiscovery}).

\subsection{Learning Architecture}\label{subsec:architecture}
Our learning architecture exploits two input streams as illustrated in Fig.~\ref{fig:main}. As with conventional MAEs, the backbone of our learning architectures is a ViT-based autoencoder, which is trained to reconstruct an input image based on a masked version of the same image. To achieve this, the autoencoder first encodes the input to a latent representation, which is then fed into a decoder to perform the reconstruction process. To feed the input into the transformer, we first divide it into non-overlapping patches and apply patch-based masking, before the remaining visible patches are fed into the encoder. In the latent space, the masked patches are considered again and the original amount of tokens is fed into the decoder for reconstruction. The reconstruction loss is computed on the same per-patch basis, as the masking is obtained. To guide the reconstruction loss towards the objects of interest, our architecture facilitates an additional stream for unsupervised OD. Within this stream, a patch-based attention map, focusing on relevant objects, is obtained from the input image. As we want to use this attention map to steer the reconstruction loss, it is important to obtain patches with the same sizes, as the MAE backbone processes. 

\subsection{Attention-Guided Loss}\label{subsec:loss}
In order to exploit the attention map obtained in our OD stream, we need to define a guiding mechanism for the model. While directly applying the attention map $\mathcal{M}$ to generate an object-based masking on the input image might seem intuitive, He~\etal~\cite{he2022masked} show that a high masking ratio in combination with random sampling significantly outperforms a block-wise masking strategy as proposed by Bao ~\etal~\cite{bao2021beit} with a lower masking ratio. Since the latter is similar to masking objects of interest in the input image, simple object-based patch masking cannot be assumed to be a promising strategy. Additionally, masking the object in the input image over and over again will not lead to masking pattern variations, which have been proven to be essential for robust MAE training. Therefore, we opt to not interfere with the input masking strategy, and rather steer the learning process by using the attention maps to guide the reconstruction loss. With this approach, we expect the autoencoder to put increased emphasis on relevant objects, without affecting beneficial patch sampling strategies.


To employ a guided loss instead of the vanilla MAE reconstruction loss, we compute the mean squared error (MSE) between the reconstructed and the normalized original image patches in pixel space, for which we only consider masked patches. With this, we obtain the per-patch reconstruction loss, i.e., each patch has its individual loss value. Having a loss value per patch, enables us to not only have a general influence on the learning process, but to guide the model in a more targeted way by individually weighting different parts of the image.


Therefore, to make our attention maps effective for guidance, they are first normalized to $\mathcal{M}_\text{norm} \in [0, 1]$. With this, all values for patches that show part of the object are close to 1 while background patches are close, but never equal, to $0$. Using the map in this form would not be useful, since background patches with values of $0$ would lead to our attention maps zeroing-out all loss values for background patches in the per-patch reconstruction loss, effectively rendering the model unable to learn to reconstruct the entire image. Our goal is instead, to keep loss values of background patches, but in a non-scaled manner. To achieve this, we scale our attention maps with the exponential function and a temperature parameter $\tau > 0$:


\begin{equation}
  \mathcal{M}_\text{scaled} = \exp\left(\frac{\mathcal{M}_\text{norm}}{\tau}\right) 
  \label{eq:att_scaled}
\end{equation}

The exponential function provides the ideal properties since background patches with values of $0$ are now converted to $1$, so that, when weighting the reconstruction, loss values for background patches are left unchanged. At the same time, the attention map values for object patches are scaled, giving them a stronger impact when weighting the loss. By adding a temperature parameter $\tau$, we obtain an additional hyperparameter, which gives us fine-grained control over the strength of weighting. 

With this, we arrive at our guided loss

\begin{equation}
  \mathcal{L}_\text{AttG} = \gamma \cdot \mathcal{L}_\text{Rec} \cdot \mathcal{M}_\text{scaled}
  \label{eq:loss}
\end{equation}

\noindent with $\gamma$ being the binary mask of all masked tokens and $\mathcal{L}_\text{Rec}$ the reconstruction loss. We schedule $\tau$ to increase according a half-cycle cosine temperature schedule starting at $\tau = 0.75$, which we find most effective as detailed in Section~\ref{sec:temperature}. Our full training process is detailed in Algorithm \ref{alg:guidance}.
As mentioned before, the temperature parameter enables us to control how strongly the MAE is guided. By decreasing $\tau$, the scaling effect on the attention map is increased, therefore providing a stronger guidance to the reconstruction loss. On the other hand, increasing it leads to a softer guidance and less enforced emphasis on reconstructing the relevant object in the scene. 
%

\begin{algorithm}[thb]
\caption{Pseudocode Attention Map Guidance}\label{alg:guidance}
\begin{algorithmic}
\Require image $\mathcal{X}$, attention map $\mathcal{M}$, \\
    masking function \textbf{mask}, model \textbf{MAE}
\ForAll{$\mathcal{X}$, $\mathcal{M}$ in Dataset}
    \State $\mathcal{X}_\text{masked}, \gamma \gets \textbf{mask}(\mathcal{X}, \text{ratio}=0.75)$
    \State $\hat{\mathcal{X}} \gets{} \textbf{MAE}(\mathcal{X}_\text{masked})$ 
    \State $\mathcal{M}_\text{norm} \gets (\mathcal{M} - \mathcal{M}_{\text{min}}) / (\mathcal{M}_{\text{max}} - \mathcal{M}_{\text{min}})$
    \State $\mathcal{M}_\text{scaled} \gets \exp{(\frac{\mathcal{M}_\text{norm}}{\tau})}$     \text{\hspace{3cm}(Eq.~\ref{eq:att_scaled})}
    \State $\mathcal{X}_{\text{norm}} \gets (\mathcal{X} - \mu_{\mathcal{X}}) / (\sigma_{\mathcal{X}} + \epsilon)^{0.5}$
    \State $\mathcal{L}_\text{Rec} \gets (\hat{\mathcal{X}} - \mathcal{X}_{\text{norm}})^2$
    \State $\mathcal{L}_\text{Guided} \gets \gamma  \cdot \mathcal{L}_\text{Rec} \cdot \mathcal{M}_{\text{scaled}}$    \text{\hspace{2.35cm}(Eq.~\ref{eq:loss})}
\EndFor
\end{algorithmic}
\end{algorithm}

\begin{figure}[thb]
\centering
\begin{subfigure}{0.247\linewidth}
    \includegraphics[width=\textwidth]{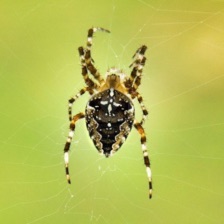}
\end{subfigure}%
\hfill
\begin{subfigure}{0.247\linewidth}
    \includegraphics[width=\textwidth]{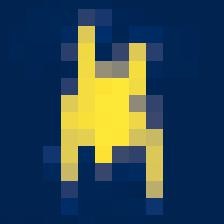}
\end{subfigure}%
\hfill
\begin{subfigure}{0.247\linewidth}
    \includegraphics[width=\textwidth]{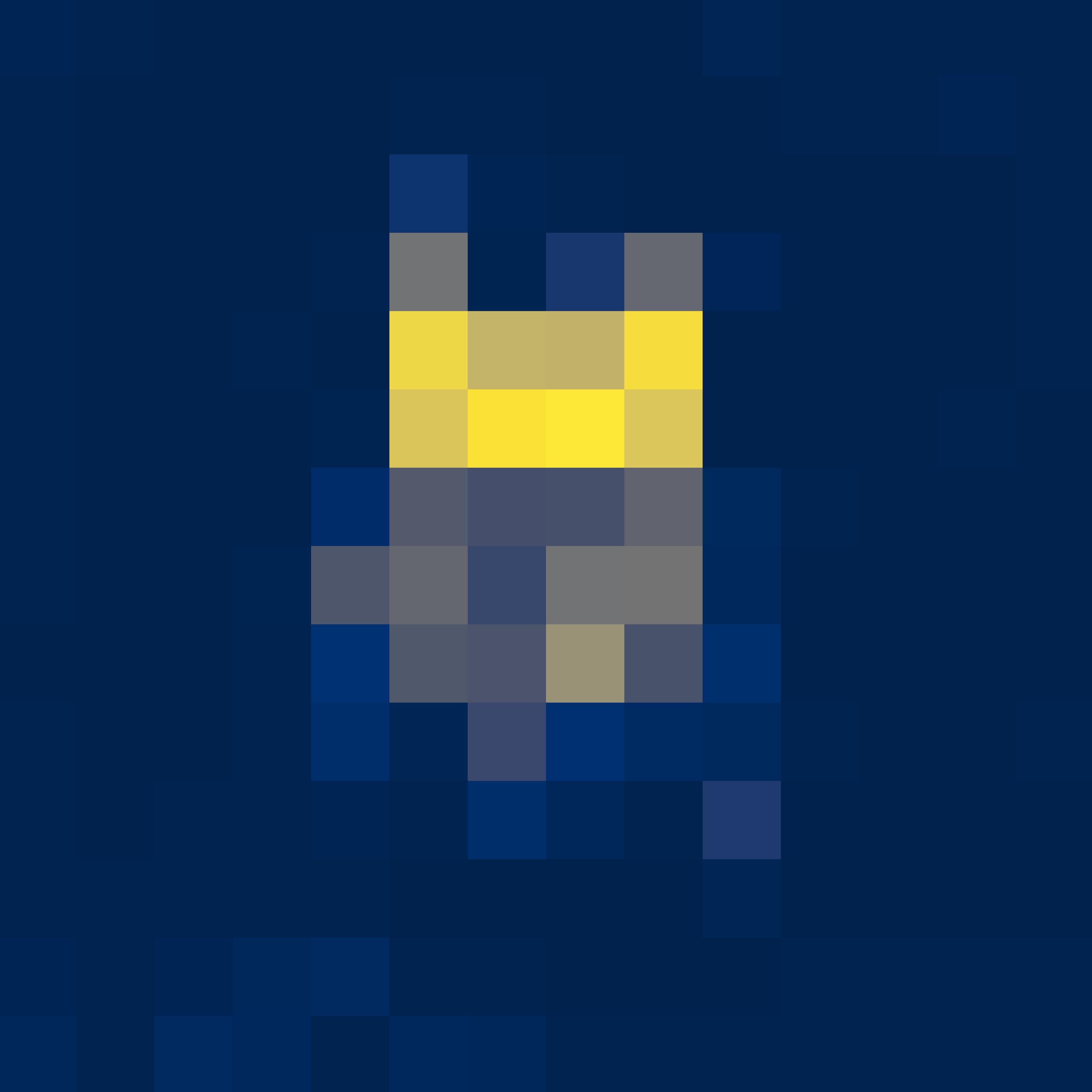}
\end{subfigure}%
\hfill
\begin{subfigure}{0.247\linewidth}
    \includegraphics[width=\textwidth]{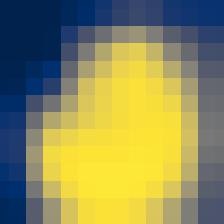}
\end{subfigure}%
\\
\begin{subfigure}{0.247\linewidth}
    \includegraphics[width=\textwidth]{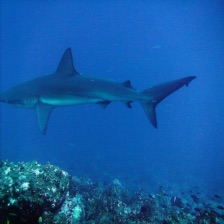}
\end{subfigure}%
\hfill
\begin{subfigure}{0.247\linewidth}
    \includegraphics[width=\textwidth]{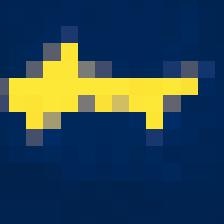}
\end{subfigure}%
\hfill
\begin{subfigure}{0.247\linewidth}
    \includegraphics[width=\textwidth]{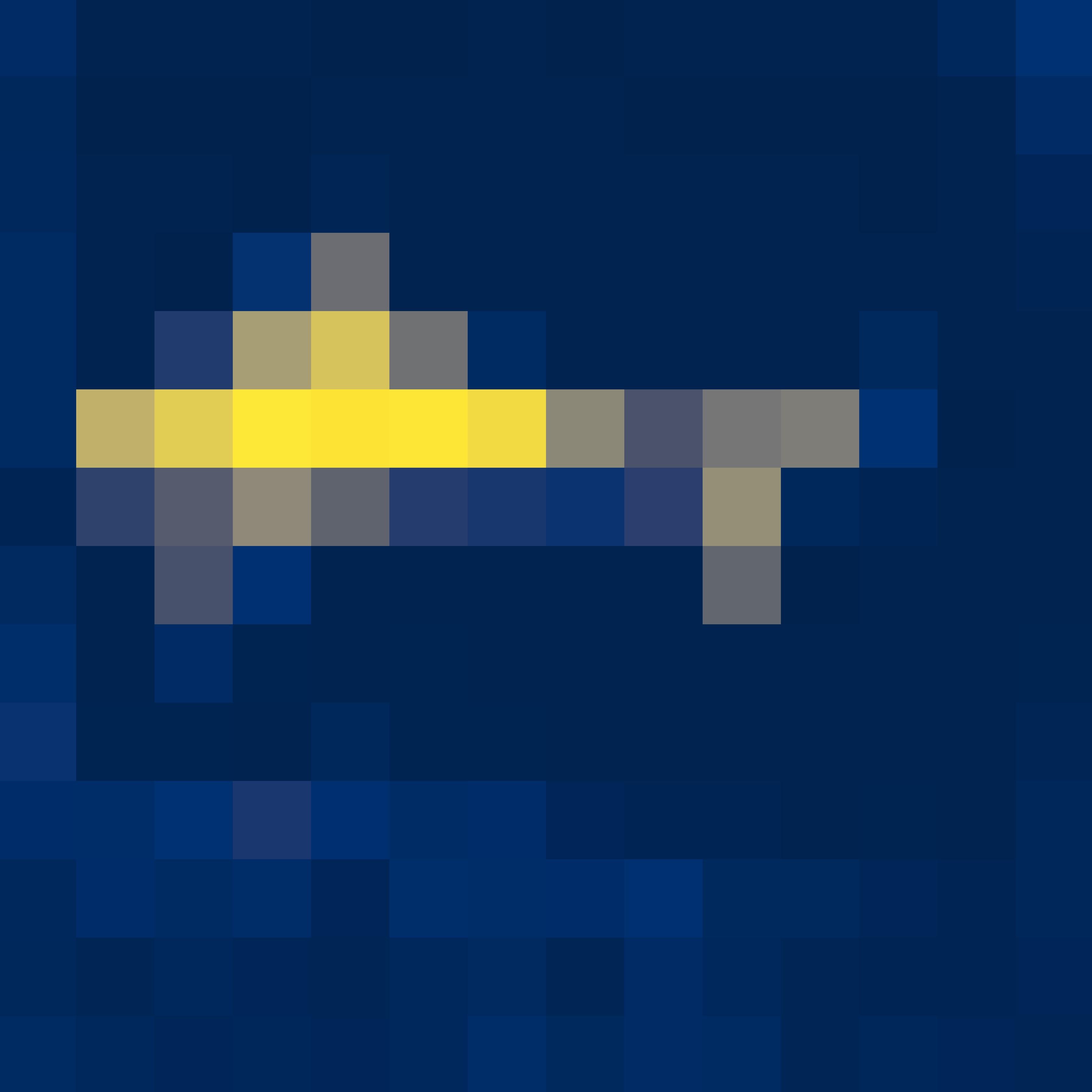}
\end{subfigure}%
\hfill
\begin{subfigure}{0.247\linewidth}
    \includegraphics[width=\textwidth]{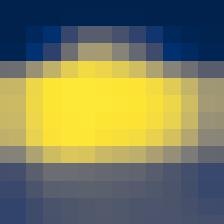}
\end{subfigure}%
\\
\begin{subfigure}{0.247\linewidth}
    \includegraphics[width=\textwidth]{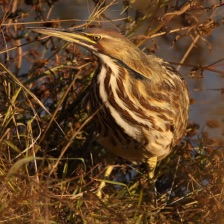}
\end{subfigure}%
\hfill
\begin{subfigure}{0.247\linewidth}
    \includegraphics[width=\textwidth]{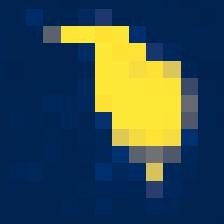}
\end{subfigure}%
\hfill
\begin{subfigure}{0.247\linewidth}
    \includegraphics[width=\textwidth]{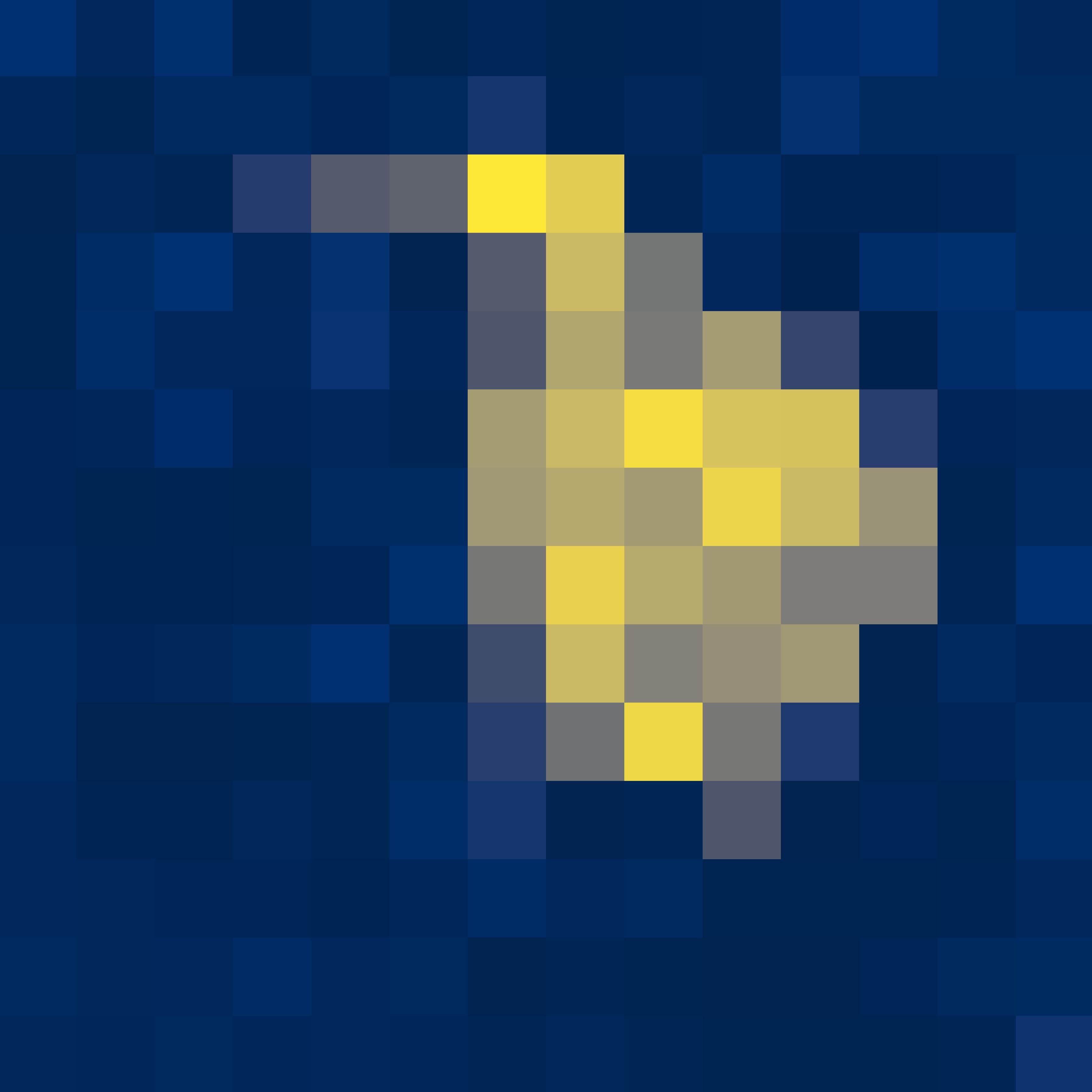}
\end{subfigure}%
\hfill
\begin{subfigure}{0.247\linewidth}
    \includegraphics[width=\textwidth]{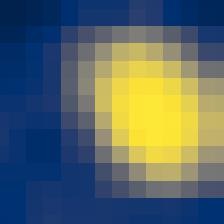}
\end{subfigure}%
\\
\begin{subfigure}{0.247\linewidth}
    \includegraphics[width=\textwidth]{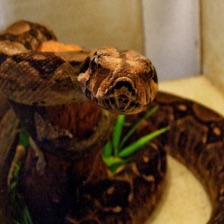}
    \caption{Original}
\end{subfigure}%
\hfill
\begin{subfigure}{0.247\linewidth}
    \includegraphics[width=\textwidth]{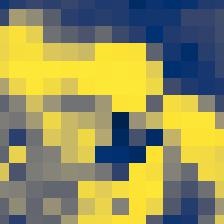}
    \caption{TokenCut}
\end{subfigure}%
\hfill
\begin{subfigure}{0.247\linewidth}
    \includegraphics[width=\textwidth]{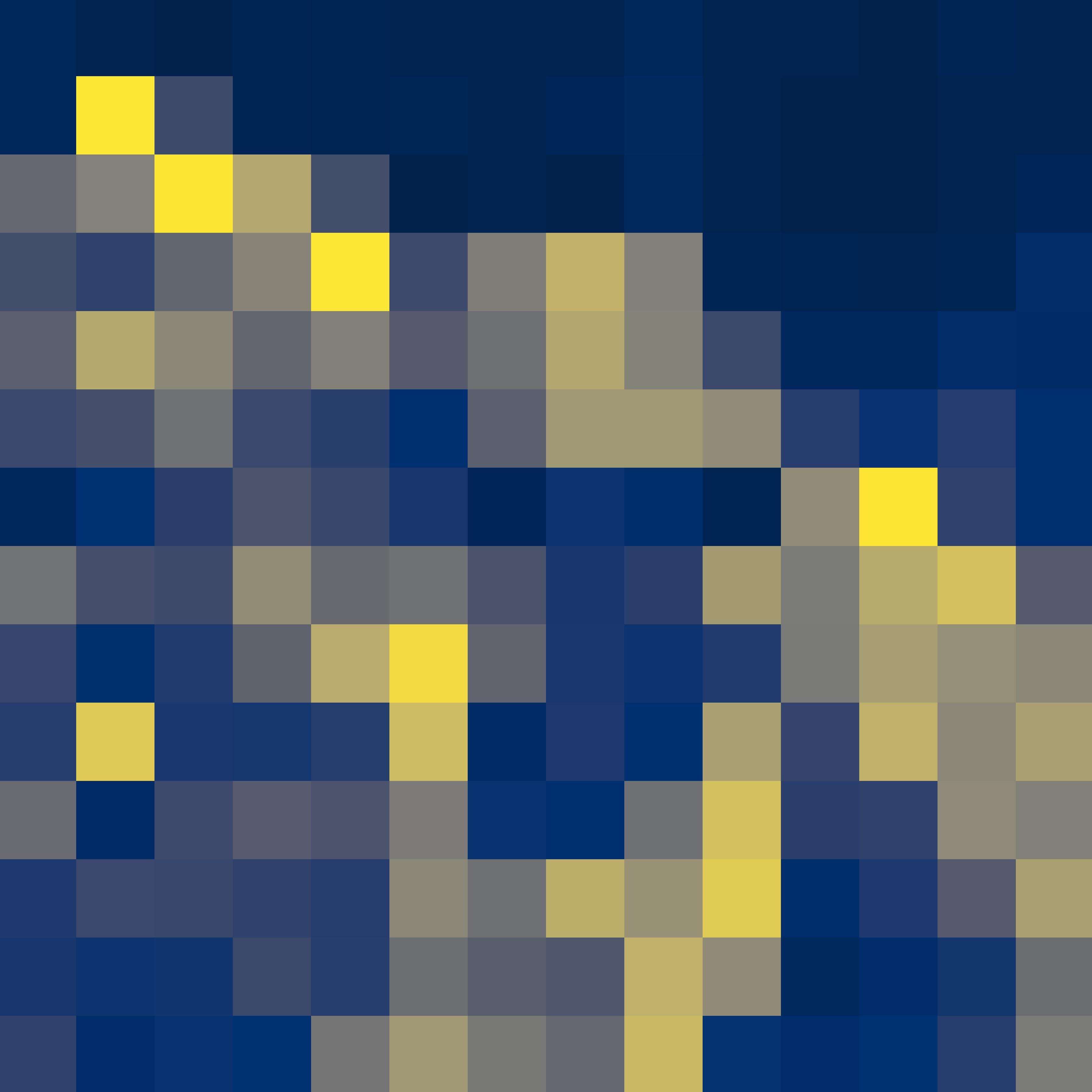}
    \caption{DINO}
\end{subfigure}%
\hfill
\begin{subfigure}{0.247\linewidth}
    \includegraphics[width=\textwidth]{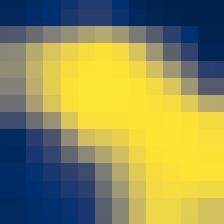}
    \caption{Grad-CAM}
\end{subfigure}%
\caption{\textbf{Visual Comparison Of Our Attention Maps.} We illustrate the differences between our attention maps. Each has been normalized and scaled in the manner as we use them for pre-training our models. From these examples, TokenCut can be perceived to output the best foreground-background separation maps of all presented methods. We study the effect of the different maps on our guidance in Section \ref{sec:mask-quality}.}
\label{fig:attention-maps}
\vspace{-4mm}
\end{figure}


\subsection{Object Discovery}\label{subsec:objectdiscovery}
The effectiveness of our guidance loss introduced above, stands and falls with the quality of the exploited attention maps. 
We ensure wide applicability of our method by exploiting state-of-the-art unsupervised OD approaches. Based on their reported results as well as their wide-spread use, we have investigated three unsupervised OD approaches to inform our guided loss: a DINO-pretrained ViT~\cite{caron2021emerging}, a TokenCut ViT~\cite{wang2022tokencut}, and a Grad-CAM-based approach~\cite{selvaraju2017grad}. 
While the first two are transformer-based approaches, which directly deliver attention maps, Grad-CAM is often praised for its capabilities to extract structures highly relevant for a learner, and has therefore also been considered. With these three approaches, we generate an attention map $\mathcal{M}$ of the object in the scene that serves as a foreground-background segmentation.


For the pre-trained DINO~\cite{caron2021emerging} ViT, the attention map is obtained, by feeding the image through the network and querying the self-attention module of the last block. This self-attention module with $H$ heads and consequently $j=1, ...,H$ sequences takes as input the normalized output sequence from the previous block $Y~\in~\mathbb{R}^{(n+1) \times d}$ and feeds it through three linear layers to map it to the sequences for query $Q_j$, key $K_j$ and value $V_j$, all $\in~\mathbb{R}^{(n+1) \times d'}$. The attention matrix is then calculated as follows:

\begin{equation}
  \mathcal{A}_j = \text{softmax}(Q_j K^T_j / \sqrt{d'})
\end{equation}

\noindent with $d'=d/H$, allowing for the maps of all attention heads to be retrieved. For DINO, we use the attention maps without post processing. 

TokenCut ViT~\cite{wang2022tokencut} post-processes the segmentation-like attention maps $\mathcal{A}_j$ of the pre-trained DINO transformer to obtain a more refined segmentation. Briefly described, TokenCut constructs an undirected graph of patches with each patch having assigned a feature vector. Each patch is linked to neighboring patches through edges, represented by the cosine similarity between the two vectors. The normalized cuts algorithm~\cite{shi2000normalized} is then used to obtain the final segmentation output $\mathcal{M}$.

The Grad-CAM-based approach~\cite{selvaraju2017grad}
visualizes the gradients for a certain prediction.
The result is a segmentation-like attention map. Differently thought, this approach originally requires labels to inform the process. Since we do not have any labels available, we select the highest scoring category as our target. Since the Grad-CAM model will output a pixel-wise heatmap, we apply average pooling to obtain patches of the same patch-wise format as our MAE and the other two ViTs. In our approach we use a supervised ImageNet-1K-pretrained ResNet-50 as the encoder.

Figure~\ref{fig:attention-maps} shows a visual comparison of the segmentation outputs. From qualitative observations, TokenCut provides the most semantically accurate attention maps with evenly valued patches and a clear distinction between object and background. DINO's maps also provide a useful semantic layout of the scene but are lacking the same distinctiveness. Grad-CAM on the other hand outputs heatmap-like masks that provides a coarser map. 
In our ablations (Section~\ref{sec:mask-quality}), we explore the effectiveness of these different maps for our guidance mechanism.



\section{Experiments}


We pre-train our model on ImageNet~\cite{deng2009imagenet} (IN1K). Our attention maps are generated by TokenCut, as they perform the best according to our ablations in Section~\ref{sec:mask-quality}. We then evaluate our learned representations with linear probing, and k-NN classification studies, both on IN1K and other relevant benchmarks, as well as low-shot finetuning and image retrieval studies.

\subsection{ImageNet Evaluation}

\begin{table}
  \centering
  \setlength{\tabcolsep}{6pt}
  \begin{tabular}{llllll}
    \toprule
    Method & Size & Epochs & k-NN & Linear \\
    \midrule
    SimMIM  & ViT-B/16 & 800 & 9.4 & 56.7 \\
    SemMAE & ViT-B/16 & 800 & 45.1 & 65.0 \\
    BEiT & ViT-L/16 & 800 & 11.4 & 73.5 \\
    \midrule
    MAE & ViT-L/16 & 800 & 52.8 & 73.5 \\
    \emph{+ AttG (Ours)} & ViT-L/16 & 800 & \underline{56.2} & \underline{74.4} \\
    \midrule
    MAE & ViT-L/16 & 1600 & 50.9 & 75.1 \\
    \emph{+ AttG (Ours)} & ViT-L/16 & 1600 & \textbf{59.0} & \textbf{75.9} \\
    \bottomrule
  \end{tabular}
  \caption{\textbf{$\bm{k}$-NN classification and linear probing on IN1K for purely MIM-based  techniques.} We show results for linear probing and $k$-NN classification, for IN1K we use $k$=20. For both training durations, our method improves upon the results of the MAE for both $k$-NN and linear probing.
  }
  \label{tab:main_results}
  \vspace{-4mm}
\end{table}

\begin{table}
  \centering
  \setlength{\tabcolsep}{6pt}
  \begin{tabular}{lllcc}
    \toprule
    Method & Size & Epochs & 1\% & 10\%  \\
    \midrule
    SimMIM  & ViT-B/16 & 800 & 23.2 & 66.3  \\
    SemMAE & ViT-B/16 & 800 & 52.8 & 70.6 \\
    BEiT & ViT-L/16 & 800  & 38.3 & 71.6 \\
    \midrule
    MAE & ViT-L/16 & 1600 & 67.1 & \textbf{78.5} \\
    \emph{+ AttG (Ours)} & ViT-L/16 & 1600 & \textbf{67.5} & 78.1 \\
    \bottomrule
  \end{tabular}
  \caption{\textbf{Low-shot finetuning on IN1K.} We finetune the pre-trained models with 1\% and 10\% of IN1K training data and evaluate on the full validation set. Our model performs especially well in the 1\% regime.
  }
  \label{tab:low-shot}
  \vspace{-3mm}
\end{table}
We collect several comparable methods in the field and report k-NN classification and linear probing results on ImageNet-1K in Table~\ref{tab:main_results}. For a leveled comparison, Table~\ref{tab:main_results} evaluates the vanilla MAE and our attention guidance method, AttG, against other pure MIM-based approaches without a contrastive objective like SimMIM~\cite{xie2022simmim}, BEiT~\cite{bao2021beit} and SemMAE~\cite{li2022semmae}. 
We show that our method is able to outperform other purely MIM-based techniques. With our attention guidance, our model improves upon the vanilla MAE on top-1 accuracy for linear probing by \textbf{+0.9\%} after 800 pre-training epochs and \textbf{+0.8\%} after 1600 epochs, showcasing the effectiveness of our learned off-the-shelf representations.
 We also perform $k$-NN classification with $k$=20 on ImageNet-1K. At 1600 epochs, we are able to outperform the MAE baseline by \textbf{+8.1\%}. 
 We also evaluate our model on low-shot finetuning with 1\% and 10\% of training data and report the results in Table \ref{tab:low-shot}. We see that our model performs best in for the 1\% subset.

Additionally, we experiment with k-NN classification in a few-shot setting with the 1600-epoch models. 
We evaluate our model with 1, 5, 10 and 20 training images on the full validation dataset. Our results in Table~\ref{tab:knn_small} show our model outperforming for all setups, while the delta increases from \textbf{+4.4\%} to \textbf{+13.5\%} with an increasing number of training images, which underlines the quality of our representations.
We also evaluate on IN1K in a few-shot linear probing setting. As presented in Table \ref{tab:few-shot}, our attention guidance improves the results of the MAE across all subsets.

\begin{table}
  \centering
  \setlength{\tabcolsep}{6pt}
  \begin{tabular}{llcccc}
    \toprule
    Method & \hspace{10mm} \# & 1 & 5 & 10 & 20 \\ 
    \midrule
    MAE & & 6.5 & 12.1 & 15.6 & 20.2 \\
    \emph{+ AttG (Ours)} & & \textbf{10.9} & \textbf{19.8} & \textbf{26.8} & \textbf{33.7}  \\
    \bottomrule
  \end{tabular}
  \caption{\textbf{$\bm{k}$-NN classification with few embedded training examples.}  We report $k$-NN classification top-1 accuracy for $k$=5 with an IN1K subset of $\# \in \{1, 5, 10,20 \}$ images. With an increasing number of training images, the delta by which our model outperforms the vanilla MAE increases. Both models have been pretrained for 1600 epochs.}
  \label{tab:knn_small}
  \vspace{-3mm}
\end{table}

\subsection{Comparison to Attention-Masking Add-On Methods}
We also compare our method against other approaches that use attention maps to build upon a pre-existing baseline, such as AttnMask~\cite{whattohide2022}, an extension of iBOT~\cite{zhou2021ibot}, and SemMAE~\cite{li2022semmae}, which also builds upon the vanilla MAE. Since these methods build upon different baselines, and therefore also start at varying levels of performance, we attempt to isolate their effects by reporting the delta in performance these methods exhibit versus their respective baselines. Therefore, in Table~ \ref{tab:add-ons}, we report the gains of AttnMask compared to iBOT, whereas for SemMAE and our method, we report additional performance versus the MAE baseline. We show that our method leads to the overall greatest improvement on k-NN performance. While the performance gains for linear probing are below those of AttnMask, our method for guidance with attention maps is the most effective for MAEs.



\subsection{Robustness to Background Changes}
ImageNet-9~\cite{xiao2020noise} (IN-9) provides a suitable dataset for evaluating robustness against background variations in image classification. It includes validation datasets for nine classes from ImageNet-1K~\cite{deng2009imagenet} where either the background or the object in the foreground have been altered, resulting in eight different benchmarks. We perform linear probing on the larger training set, ImageNet-9L (IN-9L). IN-9 contains four validation subsets which are suitable for evaluating a model's robustness to background changes: Only-Foreground (OF), Mixed-Next (MN), Mixed-Rand (MR), and Mixed-Same (MS). These subsets contain images with modified backgrounds, of which MN and MR contain out-of-distribution (OOD) backgrounds. MS is edited with a different in-distribution backgrounds, while OF only has the background removed. We report top-1 accuracy for linear probing in Table~\ref{tab:IN9_results}. Our model is able to outperform the MAE on all four tasks.

\subsection{Transfer to Other Datasets}
To evaluate the capability of our model to extract relevant features from not only ImageNet, but also other benchmarks, we perform k-NN classification and linear probing on the CIFAR-100~\cite{krizhevsky2009cifar}, Oxford Flowers~\cite{Nilsback08flowers}, and Stanford Cars~\cite{KrauseStarkDengFei-Fei_cars} datasets.
We present our results in Table~\ref{tab:transfer}. Across all tasks, our model outperforms the vanilla MAE. This yet again shows our model's capability to extract valuable features, even for datasets other our pre-training dataset. We especially see strong results on CIFAR-100, where the vanilla MAE is outperformed by \textbf{+11.5\%} on k-NN classification accuracy and \textbf{+2.3\%} for linear probing.






\begin{table}[t]
  \centering
  \setlength{\tabcolsep}{6pt}
  \begin{tabular}{llllll}
    \toprule
    Baseline & Add-On & Epochs & k-NN & Linear \\
    \midrule
    \multirow{1}{*}{\rotatebox[origin=c]{0}{iBOT}} & \emph{+ AttnMask} & 100 & +1.3 & \textbf{+1.7} \\
    \midrule
    \multirow{3}{*}{\rotatebox[origin=c]{0}{MAE}} & \emph{+ SemMAE} & 800 & - & +0.7 \\
    & \emph{+ AttG (Ours)} & 800 & +3.4 & \underline{+0.9} \\
    & \emph{+ AttG (Ours)} & 1600 & \textbf{+8.1} & +0.7 \\
    \bottomrule
  \end{tabular}
  \caption{\textbf{Comparison of performance deltas from attention-informed approaches that are built on top of different baselines.} We compare our method to SemMAE, another method that builds upon MAE, and AttnMask, a technique on top of iBOT. We report gains for linear probing and k-NN on IN1K. 
  }
  \label{tab:add-ons}
  \vspace{-3mm}
\end{table}

\begin{table}[t]
  \centering
  \begin{tabular}{p{1.5cm}*{4}{C{1.2cm}}}
  & \includegraphics[width=1.40cm]{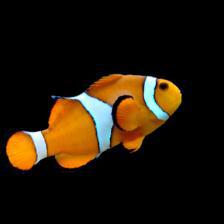} &\includegraphics[width=1.40cm]{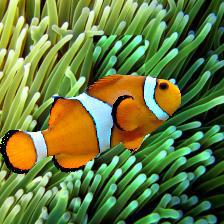} &\includegraphics[width=1.40cm]{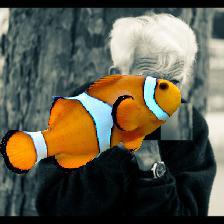} &\includegraphics[width=1.40cm]{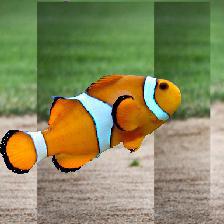}  \\
    \toprule
    Method & OF & MS & MR & MN\\
    \midrule
    MAE & 93.6 & 92.8 & 81.9 & 78.9  \\
    \emph{+AttG(Ours)} & \textbf{93.9} & \textbf{93.9} & \textbf{83.9} & \textbf{81.1} \\
    \bottomrule
  \end{tabular}
  \caption{\textbf{Background influence evaluation on ImageNet-9~\cite{xiao2020noise}.} Our model has an edge on top-1 accuracy over the vanilla MAE on three benchmarks with modified backgrounds (OF, MR, and MN), showcasing the effectiveness of our method for pre-training robust ViTs. We perform linear probing on IN-9L.}
  \label{tab:IN9_results}
  \vspace{-3mm}
\end{table}

\begin{table}[t]
  \centering
  \resizebox{\linewidth}{!}{
  \begin{tabular}{llllll}
    \toprule
    Dataset & Method & Epochs & k-NN & Linear \\
    \midrule
    \multirow{2}{*}{CIFAR100}
    & MAE & 1600 & 48.9 & 74.1 \\
    & \emph{+ AttG (Ours)} & 1600 & \textbf{60.4} & \textbf{76.4} \\
    \midrule
    \multirow{2}{*}{Flowers}
    & MAE & 1600 & 55.0 & 72.8\\
    & \emph{+ AttG (Ours)} & 1600 & \textbf{66.2} & \textbf{75.2} \\
    \midrule
    \multirow{2}{*}{Cars}
    & MAE & 1600 & 10.3 & 24.3\\
    & \emph{+ AttG (Ours)} & 1600 & \textbf{13.2} & \textbf{29.3} \\
    \bottomrule
  \end{tabular}
  }
  \caption{\textbf{$\bm{k}$-NN classification and linear probing on different datasets.} We evaluate our method on a wide variety of transfer datasets by performing linear probing and k-NN evaluation and observe consistent performance improvements with our method. Our evaluation settings are equivalent to those in Table~\ref{tab:main_results}.
  }
  \label{tab:transfer}
  \vspace{-3mm}
\end{table}

\begin{table}[t]
  \centering
  \resizebox{\linewidth}{!}{
  \begin{tabular}{llrccc}
    \toprule
    Dataset & Method & \hspace{5mm}\# & 10 & 25 & 50 \\
    \midrule
    \multirow{2}{*}{IN1K}
    & MAE & & 39.3 & 56.0 & 62.3\\
    & \emph{+ AttG (Ours)} & & \textbf{43.3} & \textbf{57.7} & \textbf{63.1} \\
    \midrule
    \multirow{2}{*}{CIFAR-100}
    & MAE & & 33.0 & 39.6 & 52.0\\
    & \emph{+ AttG (Ours)} & & \textbf{35.4} & \textbf{42.8} & \textbf{55.7} \\
    \midrule
    \multirow{2}{*}{Cars}
    & MAE & & 11.0 & 19.3 & 23.2\\
    & \emph{+ AttG (Ours)} & & \textbf{12.9} & \textbf{23.6} & \textbf{29.4} \\
    \midrule
    \multirow{2}{*}{DTD}
    & MAE & & \textbf{46.4} & 53.6 & -*\\
    & \emph{+ AttG (Ours)} & & \textbf{46.4} & \textbf{53.6} & -*\\
    \midrule
    \multirow{2}{*}{Pets}
    & MAE & & 59.1 & 69.9 & 71.3\\
    & \emph{+ AttG (Ours)} & & \textbf{69.9} & \textbf{78.9} & \textbf{80.0} \\
    \bottomrule
  \end{tabular}
  }
  \caption{\textbf{Few-shot classification using linear probing on transfer datasets.} We report top-1 accuracy results for few-shot classification  when linear probing both models on $\# \in \{10, 25, 50 \}$ images per class. Both models have been pretrained for 1600 epochs.
  *DTD train split contains only 40 images per class.
  }
  \label{tab:few-shot}
  \vspace{-3mm}
\end{table}

\subsection{Few-Shot Transfer}
We also explore effectiveness of our learned representations in different few-shot linear probing settings. For both CIFAR-100 and Stanford Cars, we limit the amount of training images to 10, 25, and 50. To test our methods robustness against changes in texture, we also evaluate on the Describable Textures Dataset (DTD) \cite{dtd} and the Oxford-IIIT Pets dataset \cite{pets}. We perform linear probing on the reduced datasets and evaluate the models' performance on the full validation splits. As detailed in Table~\ref{tab:few-shot}, our approach is able to outperform the vanilla MAE by substantial margins throughout the entire range of subset configurations and across multiple datasets. These results demonstrate once more that our attention-guided loss is highly effective in enabling MAEs to learn better representations.

\subsection{Image Retrieval}
We evaluate our learned representations with regard to their effectiveness on the revisited~\cite{philbin2008lost} Oxford and Paris image retrieval datasets~\cite{radenovic2018revisiting}. Our evaluation protocol follows Caron~\etal~\cite{caron2021emerging}. We freeze our model and retrieve the image pairs using a $k$-NN based on the $1024$-d output representations from the network. We report the mAP for the Medium (M) and Hard (H) splits in Table~\ref{tab:retrieval}. With our representations, we are able to improve the performance of the vanilla MAE across the board. 

\section{Ablations}
In all ablations, we pre-train all models for 400 epochs on ImageNet-100.

\subsection{Attention Map Quality}
\label{sec:mask-quality}
We experiment with different approaches for generating our attention maps. An overview of qualitative differences can be observed in Figure~\ref{fig:attention-maps}. The TokenCut attention map covers many details, such as the spider legs or the shark fins. DINO on the other hand outputs an attention map mostly covering the object but not with equal emphasis. In contrast, the Grad-CAM output generously overlays the entire animal but also intersects many background patches. In Table~\ref{tab:attention-map-quality}, we report the influence of using different OD approaches for deriving the attention maps informing our reconstruction loss. All models are pre-trained with an input masking ratio of $0.75$ and a half-cycle cosine temperature schedule starting at $\tau = 0.75$. 
As expected based on the qualitative results, TokenCut maps perform best with a clear advantage over the other two OD approaches. This confirms our intuition that high-quality attention maps should yield the best results, and proves the effectiveness of our approach, since the model seems to be able to extract most valuable semantic knowledge from highest quality maps.

On the other hand, the Grad-CAM attention maps seem to be more effective than the seemingly higher-quality DINO maps. This finding could point towards the necessity of enough guidance mass, since the DINO map, while being visually perceived to be more accurate, generally seems to have less impact on the reconstruction loss. 

\begin{table}[t]
  \centering
  \setlength{\tabcolsep}{6pt}
  \begin{tabular}{lcccc}
    \toprule
    Dataset \hspace{5mm} & \multicolumn{2}{c}{\bfseries $\mathcal{R}$Ox} & \multicolumn{2}{c}{\bfseries $\mathcal{R}$Par} \\
    \cmidrule(lr){2-3}
    \cmidrule(lr){4-5}
    Method & M  & H  & M  & H  \\
    \midrule
    MAE & 13.1 & 2.3 & 24.4 & 5.2 \\
    \emph{+ AttG (Ours)} & \textbf{19.1} & \textbf{3.0} & \textbf{33.0} & \textbf{8.4}  \\
    \bottomrule
  \end{tabular}
  \caption{\textbf{Image retrieval.} We evaluate on the medium (M) and hard (H) image retrieval tasks of the Oxford and Paris datasets after pre-training IN1K. We show that our network's off-the-shelf representations enable improved performance on all tasks.}
  \label{tab:retrieval}
  \vspace{-3mm}
\end{table}

\begin{table}[t]
  \centering
  \begin{tabular}{lcc}
    \toprule
    Model \hspace{20mm} & Linear & k-NN\\
    \midrule
    TokenCut \cite{wang2022tokencut} & \textbf{77.1} & \textbf{57.1}\\
    DINO \cite{caron2021emerging} & 75.7 & 56.0\\
    Grad-CAM \cite{selvaraju2017grad} & 76.3 & 57.0 \\
    \bottomrule
  \end{tabular}
  \caption{\textbf{Comparison of different OD approaches.} The TokenCut maps prove to be most effective for our guidance mechanism. This is in accordance with our  qualitative evaluation, where TokenCut seems to provide the highest quality attention maps. 
  }
  \label{tab:attention-map-quality}
  \vspace{-3mm}
\end{table}

\begin{table}[!hbt]
  \centering
  \begin{tabular}{lll}
    \toprule
    Attention Guidance \hspace{15mm} & k-NN & Linear \\
    \midrule
    Input Masking & 39.2 & 57.1 \\
    Foreground-Only & 56.7 & 74.8 \\
    Background-Only & 43.5 & 69.2 \\
    Inverted Attention Map & 48.7 & 72.6 \\
    \emph{AttG} Reconstruction Guidance (Ours) & \textbf{57.0} & \textbf{77.1} \\
    \bottomrule
  \end{tabular}
  \caption{\textbf{Ablation of different uses of the attention map.}
  Guiding the MAE through the reconstruction loss yields better results, both for k-NN classification with $k$=5 and linear probing.}
  \label{tab:attention-guidance}
  \vspace{-3mm}
\end{table}

\subsection{Computational Overhead}
Applying our attention maps to the reconstruction of the MAE comes with only marginally increased computational costs. When pre-training our models, we observed an increase in runtime of below 1\% over the vanilla MAE. Our attention maps can be generated during training, but to ensure reproducibility of our results, we decide to generate them once ahead of pre-training and use the same for all our experiments.
Obtaining an attention map requires a single forward pass of the object discovery network over the dataset, making it a negligible computational overhead in relation to the total training time. We will provide an efficient implementation for generating all attention maps.

\subsection{Guidance Mechanism}
One alternative to our proposed attention-guided reconstruction loss would be to inform the initial \emph{Input Masking} of the input image through the attention maps \cite{li2022semmae, whattohide2022}. Therefore, we investigate a modified masking strategy, where we keep our masking ratio of 75\% by sampling the values from the attention map in a descending fashion, removing patches with the highest attention values. We assume that in most cases, this masks the relevant object as well as some additional patches. Using this approach together with the original MAE loss results in significantly inferior results across multiple of our evaluations, when compared to our proposed attention-guided reconstruction loss. 
Furthermore, we experiment with different alternatives to guiding the reconstruction loss with our attention maps. We try reconstructing \emph{Foreground-Only} by zeroing all patches with attention values below the 10\%-quantile threshold. Furthermore, to emphasize background reconstruction, we use the \emph{Inverted Attention Map}s to guide the MAE and also mask out the 10\%-quantile threshold of this inverted map to reconstruct \emph{Background-Only}. 
In Table~\ref{tab:attention-guidance}, we report the top-1 accuracy for k-NN classification and linear probing on IN-100. 
As can be seen, using our attention loss guidance outperforms the alternative approaches.

\subsection{Temperature Parameter Influence}
\label{sec:temperature}

In Table~\ref{tab:temperature-parameter}, we evaluate the influence of different temperature parameters and our half-cycle temperature schedule. 
For fixed values, we observe that $\tau = 0.8$ leads to the best linear probing results. But when slowly increasing $\tau$ from $0.8$ on, we see a drop in performance, fueling the belief that further increasing the temperature is ineffective. A decrease of $\tau$ leads to worse performance in our experiments. We achieve our best results when increasing the temperature according to a half-cycle temperature schedule, starting at $\tau = 0.75$, outperforming the model with fixed $\tau = 0.75$ by $\textbf{+0.5\%}$ top-1 accuracy for linear probing.




\section{Limitations}
As outlined in Section~\ref{sec:intro}, MAEs provide state-of-the-art results for finetuning tasks, while not being able to do so for linear evaluations. As we show that our method narrows the gap for linear evaluations, it is important to also consider its impact on finetuning tasks. Our findings show that while our method does not outperform the vanilla MAE on finetuning, our results are still competitive.
Also, our approach exhibits dependence on the object discovery network used to obtain the attention maps. 

\begin{table}[t]
  \centering
  \begin{tabular}{lcc}
    \toprule
    $\tau$ \hspace{5mm} & With Schedule & Without Schedule \\
    \midrule
    0.7 & 76.1 & 76.3  \\
    0.75 & \textbf{77.1} & 76.6 \\
    0.8 & 76.5 & 77.0 \\
    \bottomrule
  \end{tabular}
  \caption{\textbf{Experiments with different temperatures and scheduling.}  We observe the best linear probing results when $\tau = 0.75$ and is increased with a half-cycle cosine schedule. }
  \label{tab:temperature-parameter}
  \vspace{-3mm}
\end{table}

\section{Conclusion \& Discussion}
We have proposed an attention-guided reconstruction masked autoencoder, which introduce an attention-guided reconstruction loss together with attention maps. Our results have shown that our attention-guided loss yields representations that enable increased top-1 accuracy for linear probing and k-NN classification across different datasets after IN1K pre-training. Further, our models exhibit improved properties for few-shot classification and have shown to be robust against realistically modified backgrounds. The presented experiments showcase the power of reconstruction guidance to learn better representations. Thus, our method presents a significant step towards enabling MAEs to learn more powerful representations already during pre-training that improve linear evaluations and k-NN classifications.

Our approach has been designed to be transferable to future MAE frameworks since we apply our guidance to the reconstruction loss. While our guidance mechanism has shown its usefulness when pre-trained on object-centric datasets, we leave it open to future work to design further guidance approaches that extend to multi-object semantic structures as can be found in COCO~\cite{lin2014microsoft}. To apply our attention-guided reconstruction loss, one could replace TokenCut with CutLER, a recent work by Wang~et~al.~\cite{wang2023cut}. This method extends TokenCut for multi-object discovery, which would allow both guiding individual objects separately or simultaneously.

{\small
\bibliographystyle{ieee_fullname}
\bibliography{egbib}
}

\end{document}